*Original Article*

# Improving Classification of Retinal Fundus Image Using Flow Dynamics Optimized Deep Learning Methods

V. Banupriya[1], S. Anusuya[2]

[1]*Department of Computer Science and Business Systems, M.Kumarasamy College of Engineering, Karur, Tamilnadu, India*
[2]*Department of Information Technology, Saveetha School of Engineering, Saveetha Institute of Medical and Technical Sciences, SIMATS, Chennai, Tamil Nadu, India*

[1]*Corresponding Author : banucs03@gmail.com*

*Abstract -* *Diabetic Retinopathy (DR) refers to a barrier that takes place in diabetes mellitus damaging the blood vessel network present in the retina. This may endanger the subjects' vision if they have diabetes. It can take some time to perform a DR diagnosis using color fundus pictures because experienced clinicians are required to identify the tumors in the imagery used to identify the illness. Automated detection of the DR can be an extremely challenging task. Convolutional Neural Networks (CNN) are also highly effective at classifying images when applied in the present situation, particularly compared to the handmade and functionality methods employed. In order to guarantee high results, the researchers also suggested a cutting-edge CNN model that might determine the characteristics of the fundus images. The features of the CNN output were employed in various classifiers of machine learning for the proposed system. This model was later evaluated using different forms of deep learning methods and Visual Geometry Group (VGG) networks). It was done by employing the images from a generic KAGGLE dataset. Here, the River Formation Dynamics (RFD) algorithm proposed along with the FUNDNET to detect retinal fundus images has been employed. The investigation's findings demonstrated that the approach performed better than alternative approaches.*

## 1. Introduction

CNNs are being used to diagnose DR by analyzing the fundus images, thus proving their superior nature when classifying or detecting tasks. The DR is a diabetes-related issue that can cause vision loss. Injury to the retina's vascular system and elevated glucose levels are the causes of this. During the year 2000, about 171 million patients offered diagnosed with diabetes. By 2030, the figure is anticipated to reach 366 million [1]. DR has various abnormal effects on the retina, such as neovascularization, hemorrhages, macular edema, microaneurysms, and soft and hard exudates. Fig. 1 depicts the various stages of DR. The DR is divided into five stages. These are mild Non-Proliferative DR (NPDR), moderate Non-Proliferative DR, severe Non-Proliferative DR, Proliferative DR (PDR), and finally, Macular Edema (ME). Mild NPDR will be the earliest stage to advance to a proliferative DR wherein vision loss occurs, and the eye gets filled with interstitial fluids. In the prior phases, the patient is generally asymptotic. Nevertheless, with the progression of the disease, the symptoms may include large floaters, distorted vision, blind spots, blurred vision, and loss of vision. Therefore, it is essential to identify the infection in its initial phases to make a precise diagnosis, strive to cut the disease's consequences drastically, and lower the chance of vision loss.

All automated techniques used to identify retinal diseases [2] using stained fundus pictures are developed to solve inadequate aspects among conventional diagnostic approaches. These decentralized technicians objectively check the patients without depending on the physicians who use gadgets to reduce the burden on qualified specialists. All earlier techniques were based on automated DR detection, but they did have many disadvantages impeding their widespread use. As most of these methods were based on shorter datasets consisting of 500 pictures for specific clinical situations, they aim to diagnose a large-scale DR for heterogeneous fundus datasets.

The techniques generated from these datasets were not always applicable to the fundus pictures (obtained from other clinical research to employ different types of eye-opening methods, fundus cameras, and sometimes both). Also, several of these algorithms that diagnose the DR depend on the manual feature extraction that aims at explaining the number of anatomically in the fundus that is predicted, like the blood vessels or the optic disc. Although these are manually tuned and applied to the individual datasets, they are still employed in detecting DR using the fundus photos of those who could meet the prototypes' demographics.

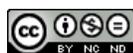

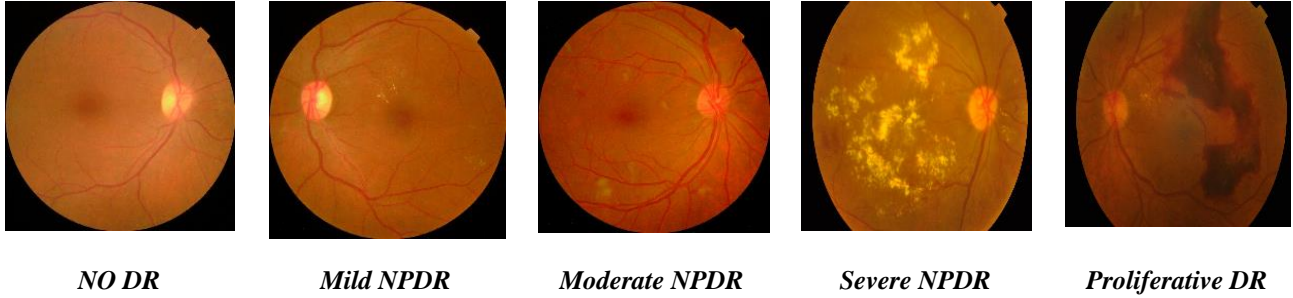

| NO DR | Mild NPDR | Moderate NPDR | Severe NPDR | Proliferative DR |

**Fig. 1 Diabetic Retinopathy Stages**

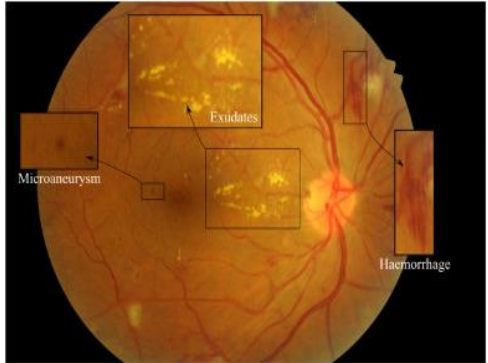

**Fig. 2 Indicative DR lesions on a fundus image**

Image classification is a crucial task in computer vision that includes several applications like the detection of objects, image segmentation [3], and localization. All methods adopted for classifying images via deep neural network, primarily CNN. The deep networks have been showing an impressive range of results. These are divided into two parts. The first one is based on CNN, which suggests the capacity of feature extraction [4]. Fig. 2 depict the Indicative DR lesions on a fundus image.

On the other hand, the second one is a fully-connected classifier that generates a new prediction model for classification. The CNN has been depicted using several hyper-parameters, especially the filter numbers, convolutional layers (CL), and their sizes. Several researchers have proposed models like Znet, AlexNet, etc. To improve the network accuracy, some decide to increase the network intensity. Even though the state-of-the-art CNNs are pretty efficient, they are designed manually.

In this research, it must be noted that misconfigured values of the CNN hyper-parameters, such as the number of filters, sizes of filters, and network depth, can affect the classifier's performance. Additionally, it may be challenging to identify the actual quantity of CLs by enumerating the use cases to choose the best possible values of the hyper-parameters. Using this work's contributions, an innovative approach known as the FUNDNET was employed to automatically include the optimal CNN hyperparameters values that result in the best CNN configuration for a particular crisis. The strategy was founded on RFD, a metaheuristic technique utilized in non-deterministic solving problems. Every CNN-based candidate structure will be encoded in the form of a graph (or gradient orientation). This method depends on "elite propagation" for the entire course of the RFD to look for the best fit for the individual. Here, deep learning-based CNNs like the FUNDNET are optimized using the RFD methods and are evaluated to classify the DR in fundus imagery. The rest of the study is structured in this way. The entire body of related material is included in Section 2. The methodologies used are explained in Section 3, the obtained measurements are covered in Section 4, and the work is wrapped in Section 5.

## 2. Related Work
Jayanthi et al. [5] developed another PSO algorithm based on the CNN Model, known as the PSO-CNN model, for detecting and classifying the DR from various color fundus prefetch images. The PSO-CNN proposed has 3 stages: preprocessing, feature extraction, and classification. In the initial stages, preprocessing is conducted as a noise removal process that can discard noise. Next, the feature extraction process that uses the PSO-CNN model was applied to the extraction of various functional feature subsets. The Decision Tree (DT) approach will then use the filtering characteristics as input to categorize the DR pictures. Simulation conducted on the PSO-CNN model used the benchmark DR database along with an experimental outcome that observed that the PSO-CNN model had completely outperformed other methods significantly. The outcome of the simulation showed that the PSO-CNN model provided maximum results.

DR detection at an early stage helped avoid any form of non-reversible blindness. Bhimavarapu and Battineni [6] incorporated fuzzy logic techniques into image processing for efficient detection. The PSO was used to partition all fundus imagery to locate microaneurysms. PSO segmentation added similarity information towards the groupings, further combining class labels. After that, a readily viewable dataset called DIARETDB0 was used for experimental validation, and a segmentation algorithm was done to use the Probability-Based (PBPSO) clustering techniques. A probabilistic continuous PSO method was used to compare the results of different fuzzy models. The outcomes demonstrated that the suggested PSO could recognize DRs earlier with an efficiency of nearly 99.9%.

Roshini et al. [7] developed an automatic DR detection model using three stages. Pre-processing consisted of two stages: converting RGB to Lab and boosting intensity. Then, the histogram equalization process was made by employing a contrast enhancement for the image. Next, progressive mean filtration was used for the filtration process, and Endurance Probability-based CSO, an enhanced meta-heuristic approach, was used to tune its constants. Lastly, the classification used the Deep CNN, in which improvement was exploited based on the convolutional layer that was further optimized in an improved FP-CSO. As all conservative CSO depended on the probability of fitness in this improved algorithm, the proposed one was called the FP-CSO. Lastly, a precious comparative performance analysis was conducted to verify the competence of the approach.

Sau and Bansal [8] presented another new DR grading. The initial process of this model was preprocessing performed using median filtering—a blood vessel with retinal abnormalities such as exudates, hemorrhages, and microaneurysms was done to grade the DR. Even before initiation is done, it was proposed to opt for optic disc removal with an open-close watershed transform. An adaptive active contour methodology will be used to identify the region of interest of the blood vessels. For this, the threshold value of the method was optimized with the FNU-GOA, which aims to increase the accuracy of DR classification. For the identified retinal abnormalities, features such as "Grey-Level Co-occurrence Matrix (GLCM)," "Local Ternary Pattern (LTP)," and "Area of Region of Interest (RoI)" were extracted. These features were subject to an MDNN used for grading the DR. The focus of the MDNN was to solve other overfitting problems, aiming to increase accuracy based on grading. The customizable proactive contour-based artery labeling was improved by the MDNN-based classification, which became increasingly reliant on the FNU-GOA. The algorithm improved grading efficiency and convergence.

Dayana and Emmanuel [31] presented another effective DNN using the Chronological Tunicate Swarm Algorithm (CTSA) to classify DR and its severity. In the initial stages, the retinal images obtained using low-quality fundus photography were preprocessed and were subject to identifying the area of interest. Firstly, the blood vasculatures were determined by using a U-Net. The area of the lesion was detected with Gabor filter banks to extract features. The classification process took place with a deep Stacked Auto-Encoder (SAE) and was jointly optimized along with a Tunicate Swarm Algorithm.

Ramya et al. [10] proposed a classification obtained from the fundus image called the hybrid CNN-based BLSO. This technique used image augmentation to increase a tissue sample to the size needed without losing any details. These characteristics first form downsized fundus images subsequently extracted as a significant feature vector and used for a feed-forward CNN. It aided the future classification of the secretions from such a fundus picture. Additionally, the PSO and BLSO were employed to optimize the hyperparameters to lessen the computational complexity. Real-time ARA400 datasets have been used to analyze research impact compared to other cutting-edge research, such as SVM classifiers, CNN, random forests, and multi-scale performance metrics. The accuracy of classification was high for this work and therefore was able to outperform the other approaches.

Bilal et al. [11] offered another new scheme for the prediction that coupled the Improved Grey Wolf Optimization (IGWO) with the CNN, known as the for diagnosing the DR Grey Wolf Optimizer (GWO). This was found to be very successful owing to the broad tuning features, good convergence speed, and scalability. This provides both exploitation and exploration for the entire search. The method uses the Genetic Algorithm (GA) to build diversified initial positions. The GWO was later applied to adjust the current population in discrete search processes and obtain an optimal feature subset in a higher classification challenge based on CNN. The technique was in contrast to the GWO, GA, and several other DR classification approaches. This strategy outperformed other methods by improving classification accuracy to 98.33%, thus proving its efficiency in detecting DR. The outcome of the simulation shows that the proposed approach could outperform other competing methods.

Pugal Priya et al. [12] proposed another deep Long LSTM used in a neural network with the RF Optimization algorithm to classify the DR. There were four different components to this: preprocessing of the image, Identifying region of interest, extraction of features and classification. To enhance the contrast ratio of the imaging, adaptive histogram equalization and histogram analysis that conducts the tissue sample and preprocess previously ignored the noise. The disease spot was also identified using an adaptable watershed based on the optic disc's size and colour. The STARE,

DRIVE, and MESSIDOR data sets were utilized for validation. The MATLAB software implemented this to several other evaluation criteria. But, the method accomplished better performance for the specificity of 98.45%, sensitivity of 96.78%, precision of 97.92%, recall of 96.89%, and F-score of 97.93% for the classification of DR.

## 3. Proposed Methodology

Rather than using a pre-trained or standard architecture, this work proposed a new sequential model, "FundNet," specifically focusing on the problem statement. The proposed model's parameters were optimized using the River Flow Dynamics-based Metaheuristic Algorithm. Kaggle datasets have about 35,000 plus training images that were graded to form five stages of the DRP, with a total of 50,000 plus test images having unidentified DRP stages. Acquiring the images was done using fundus cameras and diverse view fields. The specifics of the image acquisition, such as the sort of camera used or the angle of vision, were kept secret. Further information on the data was available on the challenge's website [35].

The selection of a subset having 1,715 images was made from the new Kaggle dataset. It has 750 images that were chosen randomly from the DRP stage 0 (no DR), 165 randomly selected images, about 450 images chosen randomly from the DRP stage 2 (moderate), about 150 images randomly selected from the DRP stage 3 (severe), and lastly, 200 images randomly chosen from the DRP stage 4 (proliferative DR). All images that show that the vision cannot be seen are excluded from the analysis of this dataset. Additionally, the chosen training images were separated into training, monitoring, and testing sets using a 60-20-20 partitioned [14].

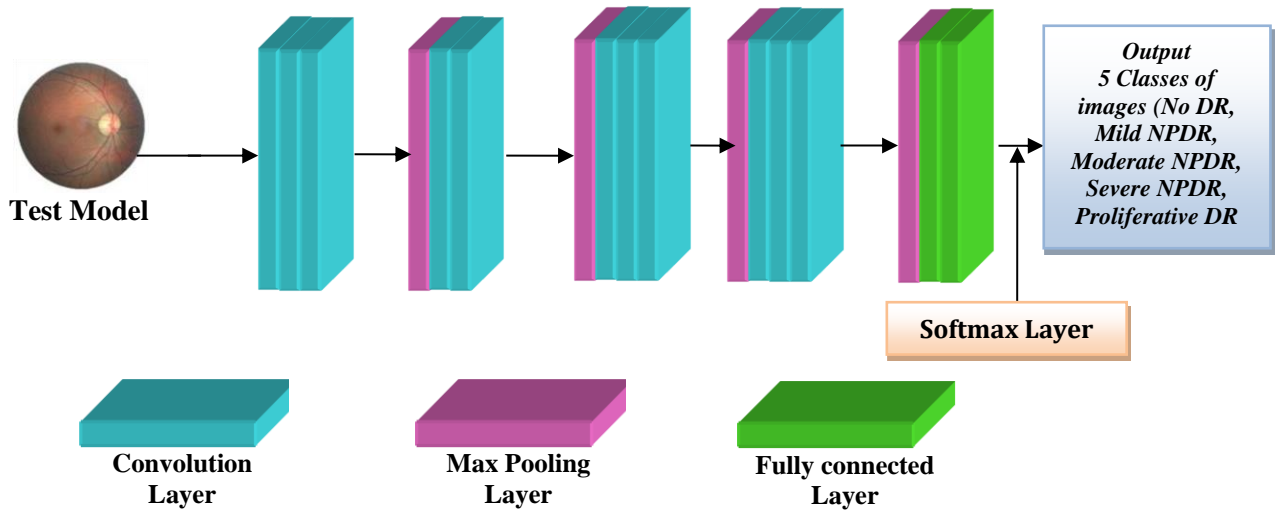

**Fig. 3 Schematic overview of proposed FundNet**

The identical patient data remained kept in the same subgroup [15]. For the preprocessing step, images were manually analyzed using random sampling to ensure and restrict the examination of the CNN to the region of interest. The input vector is made to 150000. Consequently, the input vector is too big to be used as an input node. Fig. 3 represents the schematic overview of the proposed FundNet.

The CNNs will also consider the images and their hierarchical representation when training using the stacking of multiple trainable stages. The method is state-of-the-art in image segmentation [32]. Images of three dimensions are considered based on their width, height, and the number of channels. The first two are for the resolution, while the third indicates the intensity values for the RGB colors, even if an image is converted into one dimension. The input vector is too large to be used as a network during training.

### 3.1. VGGNET, RESNET and FUNDNET

VGG-16 makes use of 3×3 filters combining them to be a sequence of the convolution for emulating the effect of amenable meadow to reduce the number of factors. VGG-16 includes 13 convolutional layers (CL), fully connected layers, and five pooling layers [17]. RESNET refers to the winner of ILSVRC 2015. This network will employ various shortcut paths performing an identity mapping to obtain the necessary output. For every residual unit, the input will be branched out where one will go to the function transformed into the "residual" branch, and the "identity" division circumvents the task. The normalization of the sample is used in the RESNET design to minimize interior cox regression shifts and speed up training [18].

Nevertheless, another issue faced by the residual net was that it could change the size of the batch, which can impact its accuracy. The accuracy will also be low for smaller batch sizes, and if the batch size is more extensive, it will require more resources. The accuracy reported used eight different GPUs consisting of 256 images. Owing to the limitation in hardware resources, they could not benchmark ResNet on various retina datasets [19-22].

The CNN architecture was employed in this work, consisting of five convolutional layers and the spatial max-pooling Rectified Linear Units (ReLUs). An ultimately linked layer plus a subsequent soft-max classifier make up the channel's concluding layers. The OxfordNet, which performed well, provided inspiration for using around 32 filters for every CL [23-26]. A max-pooling (MP) and a stride of 2 were then employed to halve the size of feature maps. By lowering the amount of design variables, MP enabled the network to become spatially invariant. With 1024 nodes and a soft-max, a completely connected layer generates a rating around 0 and 1. Also, weight decay was added, which was 5.10-5, to each layer for the penalization of outsized load factors in the back-propagation of the gradient.

### *3.2. Proposed RFD-FUNDNET Algorithm*

This work focused on optimizing the hyperparameter (with the RFD algorithm), which is a specific learning algorithm, the CNN. Also, the standard Artificial Neural Networks (ANN), along with the CNNs feature hyperparameters similar to the number of filters for each filter size, were employed. A significant challenge for this was evaluating a hyperparameter setting for CNN. It was the case for deeper models that have a higher number of filters.

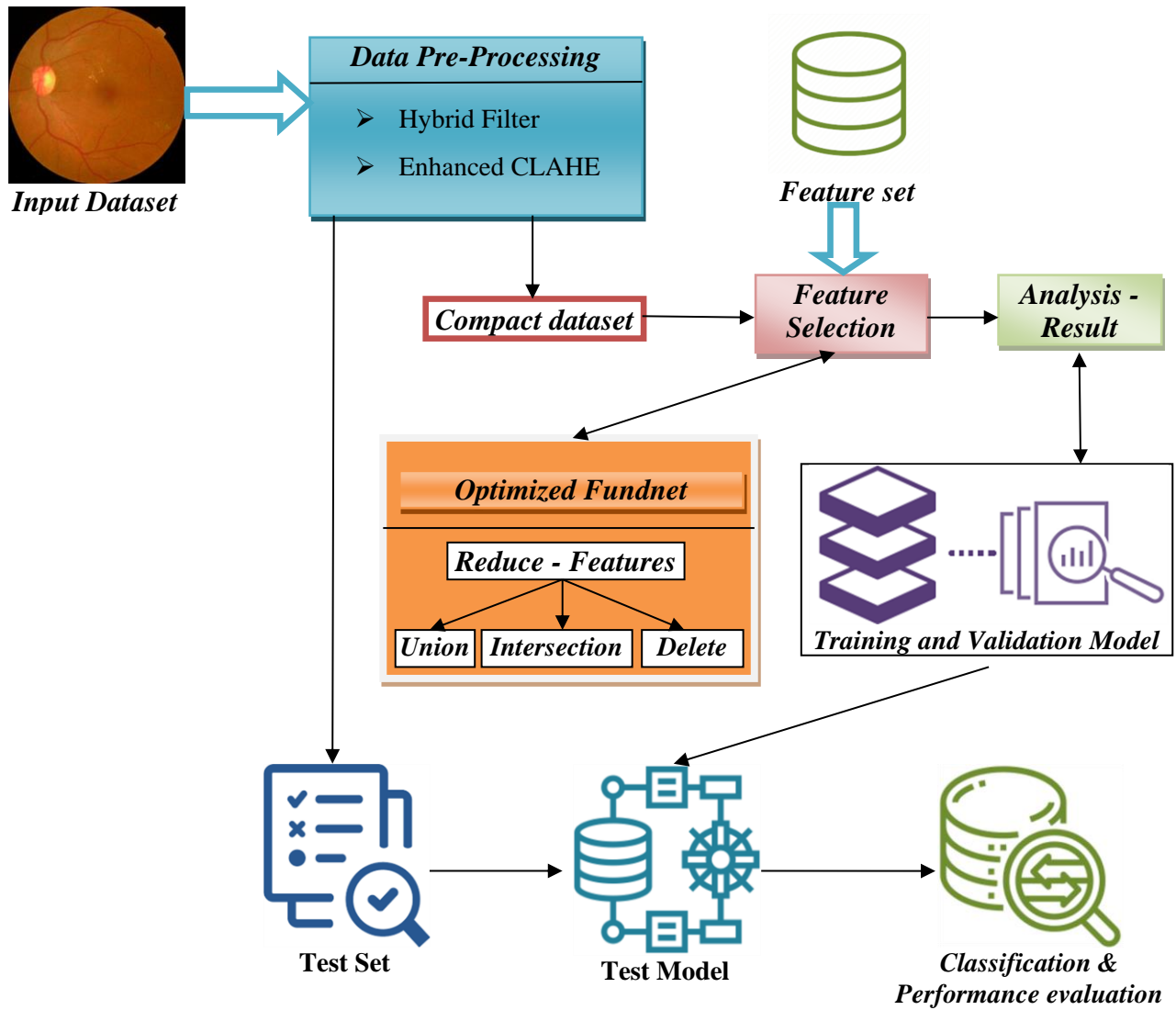

**Fig. 4 Proposed FUNDNET Framework**

The result was that the CNN inputs were simplified, or their size was brought down. But, all recent studies prove that images having higher resolution are well-suited for various tasks. It is relevant when the region of interest tends to be small, or there are deformations from the rescaling process. So, it is better to use high-resolution images though they may be impracticable for hyperparameter optimization [33]. If the hyperparameter values in images for both low and high resolution are similar, suitable hyperparameters for low-resolution ideas can be used to fine-tune them. This is very similar to the optimization of hyperparameters in various datasets. These are well-suited for a particular dataset and can be a good start for optimizing other datasets. But, hyperparameters cannot be used from different datasets as promising areas in this space should be identified based on a lower-dimensional representation of such data. Fig. 4 presents the proposed FUNDNET Framework.

The RFD was used earlier for solving several Non-Deterministic Polynomial problems in identifying paths for a graph. As the problem of finding a path for the traffic sign is brought as a graph problem, using this heuristic method was justified. Also, the algorithm had the capacity to optimize the path and its length while considering any further dependencies like vehicle energy and its consumption or time taken for travel. This is crucial during the time pathfinding is made, with acceleration and velocity being the polyhedral obstacles known as an NP-hard problem [28-30].

The RFD algorithm can be described as follows. The drops, with their movement, will erode the path or deposit any sediment. The gradient, proportionate to the fundamental gap between the node where its decrease is and its companion, will determine the probability of selecting the next node. The starting ecosystem will be uniform, with all nodes having identical heights beside the destination node. It will always be at zero. The role in establishing will receive the falls to allow for possible future research. In order to carry out erosion on the active nodes, a cluster of declines would travel the entire space in sequence for every phase. Method 1 shows the RFD procedure as a pseudo-code.

### *3.3. RFD Algorithm*

1. Nodes height <-- initial height.
2. Target node Height <-- 0.
3. While
4. end conditions - not met.
5. Do
6. Place all drops in the start node.
7. Move every drop out across the graph for the most moves possible.
8. Examine the entire path.
9. Nodes height on paths = (Nodes height on paths) - (erosion based on path costs)
10. All nodes height = All nodes height + amount of deposit
11. End while

The drops proceed one by one moment until they've passed through the required amount of nodes. This is selected based on the nodes in the surroundings. Probability $P_a$ (b, c), which is a drop in node b, will choose the subsequent node c in the following (1 to 3):

$$P_a(b,c) = \begin{cases} \frac{grad(a,b)}{total}, & for\ b\ \in\ V_a(b) \\ \frac{\vartheta|grad(a,b)|}{total}, & for\ b\ \in\ U_a(b) \\ \frac{\delta}{total}, & for\ b\ \in\ F_a(b) \end{cases} \quad (1)$$

Where in,

$$grad(b,c) = \frac{alt(b)-alt(c)}{distance(b,c)} \quad (2)$$

$$total = \left(\sum_{i \in V_a(b)} grad(b,i)\right) + \left(\sum_{i \in U_a(b)} \frac{\omega}{grad(b,i)}\right) + \left(\sum_{i \in F_a(b)} \delta\right) \quad (3)$$

$V_k$ (b) refers to neighbouring nodes having a +ve gradient, $U_a$ (b) refers to a neighbouring nodes having a -ve gradient, and $F_a$(b) has a flat gradient. The ω and δ will be some fixed values. Once the drops move, a process of erosion will be executed on the traveled route, and this is done by bringing down the node altitude based on their ascent in a succeeding node. Based on the amount of utilized drop Dp, the number of nodes in the network Ns, and an attrition factor En, the attrition for every pairing of nodes is b and c (Eq. (4)).

$$\forall b,c \in Path_a, alt(b) = alt(b) - \frac{En}{(M-1)Dp}.grad(b,c) \quad (4)$$

Wherein, Path$_a$ refers to a path go across by the drop a. In addition, if a drop stops moving, a deposit is made of a fraction of the sediment carried and will evaporate for the remaining algorithm iteration. This could also lessen the likelihood that a conversion will be performed to a blind alley, diminishing the hazardous routes [34]. A small quantity of sediment will be put into the terminals after each iteration is finished, preventing a circumstance where the elevations are virtually zero as well as the variations are insignificant, and it would follow the created courses. The equation provides the equation for both the additional sediment (5).

$$\forall b \in Grap\ ^\wedge b \neq target, alt(b) := alt(b) + \frac{erosion\ produced}{M-1} \quad (5)$$

Where Grap refers to a group of nodes, the target was the target node, and the erosion was the totting up of all erosion in progress iteration, which is

$$\sum_{Route_a} \frac{E}{(M-1)D} grad(b,c), \forall a \in drops$$

Until the last criterion has been satisfied, this algorithm was employed. This condition refers to the drops in one path. Additionally, a top limitation on the exact amount of iterations that can be supplied was included to boost performance, as well as a predicate that could be used to check whether the result had improved in the concluding n iterations.

## 4. Performance Analysis

In this section, the VGG19, RESNET50, FUNDNET, and RFD-FUNDNET processes are utilized. The results are summarized in Table 1. The sensitivity, specificity, and f measure (no DR, mild, moderate, severe, and proliferative) as exposed in Fig (5 - 7).

**Table 1. Summary of Results**

|  | **VGG19** | **RESNET 50** | **FUNDNET** | **RFD-FUNDNET** |
|---|---|---|---|---|
| Sensitivity-NO DR | 0.9227 | 0.9467 | 0.9627 | 0.9787 |
| Sensitivity-Mild | 0.8485 | 0.903 | 0.9212 | 0.9455 |
| Sensitivity-Moderate | 0.9089 | 0.9311 | 0.9533 | 0.9667 |
| Sensitivity-Severe | 0.86 | 0.9133 | 0.9267 | 0.9533 |
| Sensitivity-Proliferative DR | 0.835 | 0.905 | 0.93 | 0.95 |
| Specificity-NO DR | 0.969 | 0.9801 | 0.9848 | 0.9893 |
| Specificity-Mild | 0.9742 | 0.983 | 0.988 | 0.9908 |
| Specificity-Moderate | 0.9758 | 0.9833 | 0.9868 | 0.9919 |
| Specificity-Severe | 0.9624 | 0.9753 | 0.9841 | 0.9908 |
| Specificity-Proliferative DR | 0.9779 | 0.9868 | 0.9897 | 0.9939 |
| F Measure-NO DR | 0.9421 | 0.9608 | 0.9718 | 0.9826 |
| F Measure-Mild | 0.8187 | 0.879 | 0.9074 | 0.9313 |
| F Measure-Moderate | 0.9222 | 0.9426 | 0.9586 | 0.9721 |
| F Measure-Severe | 0.7725 | 0.8457 | 0.8882 | 0.9316 |
| FMeasure-Proliferative DR | 0.8392 | 0.905 | 0.9277 | 0.9524 |

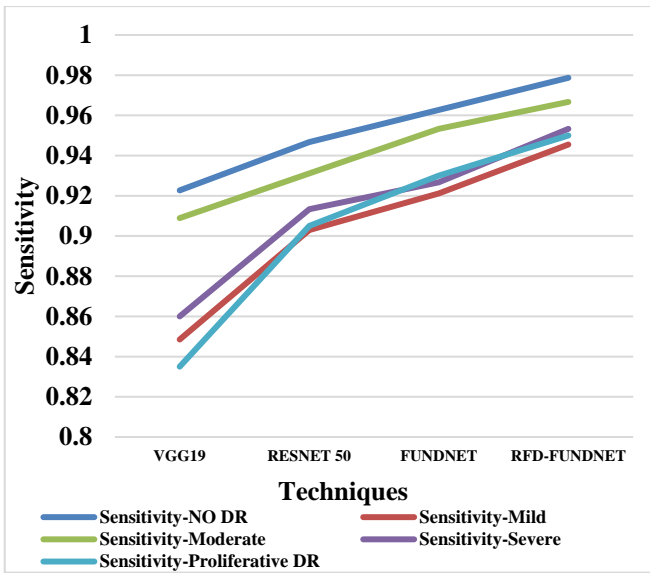

**Fig. 5 Sensitivity for RFD-FUNDNET**

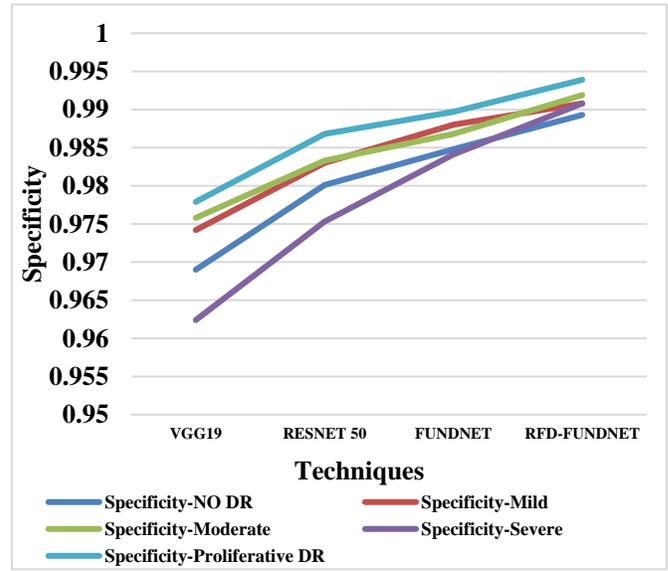

**Fig. 6 Specificity for RFD-FUNDNET**

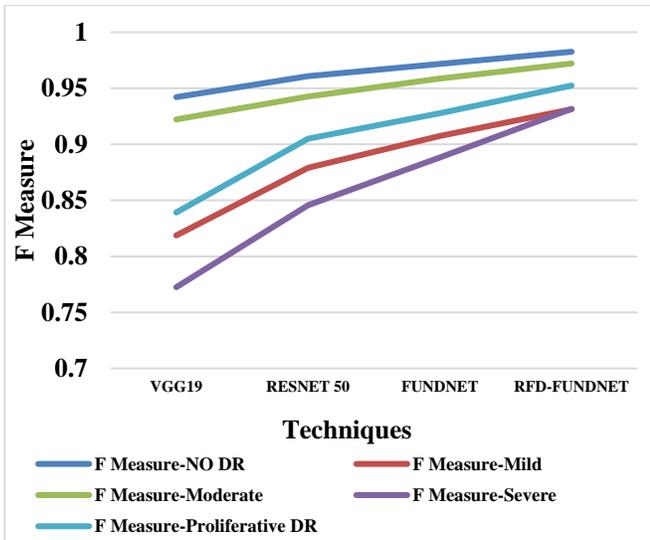

**Fig. 7 F Measure for RFD-FUNDNET**

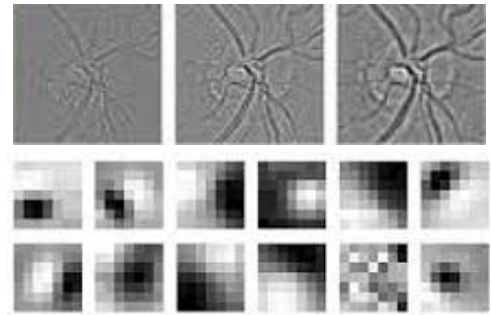

**Fig. 8 Some of the filter outputs**

Fig.5. reported that the mean sensitivities of the RFD-FUNDNET are greater than 9.14% for VGG19, 4.15% for RESNET50, and 2.11% for FUNDNET, respectively.

In Fig. 6, the RFD-FUNDNET has higher average specificity of 1.98% for VGG19, 0.97% for RESNET50, and 0.47% for FUNDNET, respectively.

In Fig. 7, the RFD-FUNDNET has a higher average f measure of 10.48% for VGG19, 5.09% for RESNET50, and 2.46% for FUNDNET, respectively. It shows the schematic computation done in the filter, and Fig. 8 shows the resultant output for the given FUNDNet Architecture.

## 5. Conclusion

Today, clinicians have been diagnosing DR by examining the lesions associated with vascular anomalies. The approach can also be very effective in terms of cost, but minute lesions are not visible in their primary stage. For this work, a productive method based on CNN had been utilized for classifying the images, and these processes are added to deep learning (such as VGGNET and RESNET) classification of the DR. For the purpose of this work, the RFD-FUNDNET has been proposed. The RFD refers to a metaheuristic wherein the solutions have been constructed by an iterative modification of the values connected to the nodes in a graph. The Gradient orientation can provide certain features like quick reinforcement of new shortcuts, elimination of blind alleys, and avoidance of cycles. The results prove that RFD-FUNDNET had a higher average sensitivity of 9.14% for the VGG19, by about 4.15% for the RESNET50, and finally, by about 2.11% for the FUNDNET, respectively.